\documentclass[a4paper, 10pt, conference]{ieeeconf}         

\usepackage{amsmath,amssymb,amsfonts}
\usepackage{graphicx}
\usepackage{pifont}

\newcommand{\DisableFootNotes}{%
  \renewcommand{\footnote}[2][]{\relax}
}

\DisableFootNotes
\IEEEoverridecommandlockouts 
\usepackage{xcolor}
\usepackage[figuresright]{rotating}

\usepackage{graphicx}
\graphicspath{{/}{fig/}}

\usepackage{array}
\usepackage{textcomp}
\usepackage{xcolor}
\usepackage{multirow}
\usepackage{booktabs}

\usepackage{mathtools}
\usepackage{breqn}
\usepackage{float}

\usepackage{pgfplots}
\pgfplotsset{compat=1.7}
\usepgfplotslibrary{groupplots}

\usepackage{caption}
\usepackage{subcaption}

\usepackage{multirow,tabularx}
\usepackage{hyperref}
\usepackage{flushend}
\usepackage{algorithmic}
\usepackage[linesnumbered, ruled, vlined]{algorithm2e}

\newlength\figureheight
\newlength\figurewidth
\SetAlCapNameFnt{\footnotesize}
\SetAlCapFnt{\footnotesize}

\usepackage[left=1.91cm,right=1.31cm,top=3.67cm,bottom=1.91cm]{geometry}

\title{
    \LARGE \bf
    Event-driven Fabric Blockchain - ROS\,2 Interface: \\
    Towards Secure and Auditable Teleoperation of Mobile Robots \\
}

\author{%
        \vspace{1em}
        Lei Fu$^{1}$,
        Salma Salimi$^{1}$,
        Jorge Pe\~na Queralta$^{1}$,
        Tomi Westerlund$^{1}$\\
        \normalsize
        $^{1}$\href{https://tiers.utu.fi}{Turku Intelligent Embedded and Robotic Systems (TIERS) Lab, University of Turku, Finland}.\\
        Emails: \textsuperscript{1}\{leifu, salmas, jopequ, tovewe\}@utu.fi\\[+6pt]
}

\begin{document}

\maketitle
\thispagestyle{empty}
\pagestyle{empty}

%%%%%%%%%%%%%%%%%%%%%%%%%%%%%%%%%%%%%%%%%%%%%%
%%                                          %%
%%           ABSTRACT AND TITLE             %%
%%                                          %%
%%%%%%%%%%%%%%%%%%%%%%%%%%%%%%%%%%%%%%%%%%%%%%

%%%%%%%%%%%%%%%%%%%%%%%%%%%%%%%%%%%%%%%%%%%%%%
%%                                          %%
%%                ABSTRACT                  %%
%%                                          %%
%%%%%%%%%%%%%%%%%%%%%%%%%%%%%%%%%%%%%%%%%%%%%%

\begin{abstract}%
    \label{sec:abstract}%
    The integration of blockchain technology in robotic systems has been met by the community with a combination of hype and skepticism. The current literature shows that there is indeed potential for more secure and trustable distributed robotic systems. However, it is still unclear in what aspects of robotics beyond high-level decision making can blockchain technology be indeed usable. This paper explores the limits of a permissioned blockchain framework, Hyperledger Fabric, for teleoperation. Remote operation of mobile robots can benefit from the auditability and security properties of a blockchain. We study the potential benefits and the main limitations of such an approach. We introduce a new design and implementation for a event-driven Fabric-ROS\,2 bridge that is able to maintain lower latencies at higher network loads than previous solutions. We also show this opens the door to more realistic use cases and applications. Our experiments with small aerial robots show latencies in the hundreds of milliseconds and simultaneous control of both a single and multi-robot system. We analyze the main trade-offs and limitations for real-world near real-time remote teleoperation.
\end{abstract}

\IEEEpeerreviewmaketitle

%%%%%%%%%%%%%%%%%%%%%%%%%%%%%%%%%%%%%%%%%%%%%%
%%                                          %%
%%                SECTIONS                  %%
%%                                          %%
%%%%%%%%%%%%%%%%%%%%%%%%%%%%%%%%%%%%%%%%%%%%%%
%%%%%%%%%%%%%%%%%%%%%%%%%%%%%%%%%%%%%%%%%%%%%%
%%                                          %%
%%              INTRODUCTION                %%
%%                                          %%
%%%%%%%%%%%%%%%%%%%%%%%%%%%%%%%%%%%%%%%%%%%%%%

\section{Introduction}\label{sec:introduction}

As robots become more ubiquitous, connected, and are deployed in larger scale, there is significant potential in integrating technologies and approaches from the domain of distributed networked systems~\cite{simoens2018internet, zhang2022distributed}. Indeed, a sizeable amount of research has been directed to the integration of blockchain technologies into robotic systems~\cite{ferrer2018blockchain, strobel2018managing, ferrer2021following}. However, there is a natural skepticism in the robotics community in terms of the scalability of these systems and their applicability to real-world use cases. Recent research is shedding light on practical applications, with novel distributed ledger technologies (DLTs) beyond linear blockchains~\cite{keramat2022partition}, as well as in permissioned blockchains that demonstrate more realistic use cases~\cite{salimi2022secure}.

One of the key properties of blockchain-based autonomous robotic systems is the ability to audit the autonomy decisions a posteriori, owing to the immutability of the data stored in the blockchain~\cite{white2019black}. The same benefit can be extended to teleoperated robots. However, current solutions fail to meet the needs of real-time teleoperation or shared autonomy~\cite{aditya2021survey}.

\begin{figure}[t]
    \centering
    \includegraphics[width=0.45\textwidth]{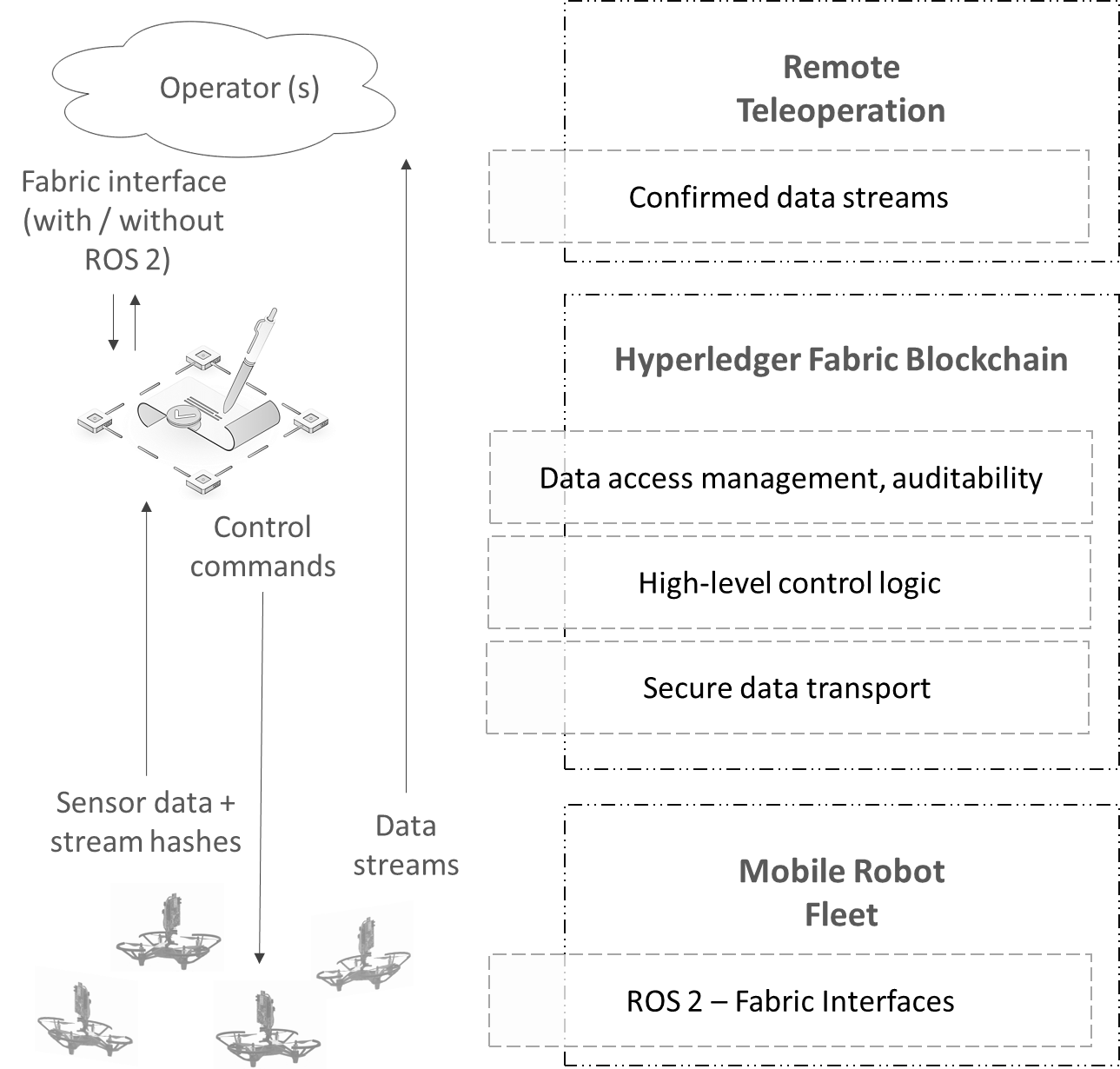}
    \caption{Conceptual illustration of the proposed approach, where a Hyperledger Fabric blockchain network bridges robots with a remote teleoperation application. The Fabric network is able to transport data with low latencies in the range of hundreds of milliseconds. Real-time data streams (e.g., live video) can be confirmed through hashes in the blockchain. The Fabric network allows for control access management and auditability.}% of the teleoperation.}
    \label{fig:concept}
    \vspace{-2em}
\end{figure}

The majority of the work today in blockchain and robotics relies on the permissionless and open Ethereum network. Smart contracts running on the Ethereum Virtual Machine have enabled various applications such as secure federated learning~\cite{ferrer2018robochain} and trustable vehicular networks~\cite{xianjia2021flsurvey}. However, Ethereum faces challenges with scalability, as well as adaptability to the needs of industrial robotic applications, even with the proof-of-stake consensus~\cite{queralta2021blockchain}. %as it relies on cryptographic proof-of-work for its consensus algorithms~\cite{queralta2021blockchain}.}

Albeit the possibility of deploying private Ethereum networks, the framework is not designed to support natively identity management and data access permissions. It is therefore not necessarily suitable for industrial applications. The Hyperledger Fabric blockchain framework, on the other hand, has been designed based on a series of principles that can drive adoption in the robotics community: identity management through certificate authorities, data access control policies and private data channels, among others. We have previously shown the potential of Fabric - ROS\,2 interfaces for data recording~\cite{salimi2022towards}, for multi-robot collaboration~\cite{salimi2022secure}, and also for cooperative decision-making processes~\cite{torrico2022UWBRole}. In the previous works, however, ROS\,2 and the blockchain have not been integrated in a way that it would allow for generic data transport (e.g., configurable ROS topics), high-level logic for data access management, or near real-time two-way data transmission. This paper therefore focuses on providing a more general, more performant, event-driven interface.

The main objective of this paper is to advance in the integration of blockchain technology in real-world robotic applications. Towards that end, the paper's advances are twofold. First, we redesign and re-implement our previous work on Fabric and ROS\,2 integration~\cite{salimi2022towards}, with an event-based approach that allows for more seamless connection to ROS, as well as better separation of logic and confirmation of data outside the blockchain. Second, we investigate the potential for applications requiring near real-time operations with two-way data transmission, in comparison with previous works in the literature featuring only non-time-critical decision making through the blockchain~\cite{salimi2022secure, mokhtar2019blockchain}. The objective is to allow for secure and auditable transmission of ROS data between different ROS\,2 networks. The closest example in the literature is~\cite{mokhtar2019blockchain}, where the authors use Hyperledger Fabric to explore a new path planning approach for multi-robot systems to achieve secure and scalable distributed control systems. However, in~\cite{mokhtar2019blockchain} the real-time factor is rather flexible, with computation and confirmation times reaching over 5\,s. Additionally, there is no standard methodology for integrating such approach with ROS or ROS\,2, and instead the authors provide a very specific implementation that is hard to generalize to other robotic use cases or application domains. In summary, the main contributions are the following:
\begin{enumerate}
    \item a novel approach to bridging and interfacing ROS\,2 systems with a Hyperledger Fabric blockchain network, through an event-based design and parametrizable data transport settings; 
    \item the design, implementation and evaluation of a proof of concept for near real-time robotic teleoperation through Fabric, from single to multi-robot systems.
\end{enumerate}

Our experimental results with both simulation and real-world experiments. Our platform is built to provide a control interface with human operators. However, automated scripts control the robots in practice to enable comparative experiments and quantitative analysis. The experiments are done with a set of micro-aerial vehicles (MAVs), with one and two drones in the real-world experiments, and up to 16 drones in simulator environments to show how the proposed system scales. The MAVs we use are DJI Tello drones, operated in a motion capture (MOCAP) system for external positioning.

The rest of this document is structured as follows. Section II examines previous research on using blockchain technology for robotics and the integration of Hyperledger Fabric and ROS. Section III introduces the high-level system design, while Section IV details the experimental setup and approach. The findings of the experiments are presented in Section V and the work concludes in Section VI.

%%%%%%%%%%%%%%%%%%%%%%%%%%%%%%%%%%%%%%%%%%%%%%
%%                                          %%
%%              RELATED WORKS               %%
%%                                          %%
%%%%%%%%%%%%%%%%%%%%%%%%%%%%%%%%%%%%%%%%%%%%%%

% \newpage
\section{Background} \label{sec:related_work}

Through this section, we discuss the existing literature at the intersection of blockchain and robotics, identifying some of the main challenges that are addressed by this paper. We also briefly introduce the key concepts behind the Hyperledger Fabric blockchain framework.

\subsection{Blockchain in Robotics}

Blockchain systems have several characteristics that can be used in distributed multi-robot systems. First, built-in security that facilitates data sharing within networked systems~\cite{ferrer2018blockchain, wang2019survey, abichandani2020secure}. Second, immutability that allows for data to be audited and tracked~\cite{white2019black}. Third, consensus protocols that support collaborative decision-making through the use of smart contracts~\cite{keramat2022partition, nguyen2019blockchain}. %These properties have been highlighted in various studies ~\cite{sankar2017survey,white2019black,wang2019survey, nguyen2019blockchain}.

Strobel et al. proposed an early proof-of-concept solution to address security challenges in swarm robotic systems~\cite{strobel2018managing}. The authors used Ethereum and decentralized smart contracts to create secure coordination mechanisms for identifying and eliminating malfunctioning members within the swarm. They aim to demonstrate the possibility of achieving this with their proposed solution.

It is clear that blockchains can provide an underlying framework for achieving consensus within a multi-robot system or robot swarm. A recurrent use case in the literature is the detection of byzantine or malicious agents in the system~\cite{ferrer2018blockchain, strobel2018managing, salimpour2022decentralized}. %There has been ongoing research into using blockchains as a way to achieve consensus among robot swarms and detect any Byzantine agents, this area of research has been actively studied, as demonstrated in various studies~\cite{ferrer2018blockchain, strobel2018managing}.
To achieve security, limited memory, and resilience to deception, Ferrer et al. in~\cite{ferrer2021following} use blockchain as an asynchronous registry for messages transmitted between leaders and robots with gregarious attributes. %It has also proposed a method that can also be used in practical scenarios where there are resource limitations for the Byzantine Follow Multiple Leaders and Byzantine Loosely Follow Multiple Leaders problems. \red{This paper also evaluated the blockchain layer's resource usage impact, which we found to be negligible in complex robotic systems. (are we going to check the resource utilization or not?)}

From the perspective of distributed networked systems, in~\cite{aloqaily2021energy} the authors propose a network solution for multi-robot systems that enhances the capabilities of both aerial drones (UAVs) and ground robots (UGVs), improves connectivity, increases service availability, and is energy-efficient. This solution also uses a service composition approach to provide user-specific services based on their requirements and utilizes a consensus algorithm and blockchain technology to ensure the integrity and authenticity of sensitive services by checking locally trained models.
% \red{In this paper we have used Fabric smart contracts for robot decision-making while updating the data in the blockchain.}

\subsection{Hyperledger Fabric}

Hyperledger Fabric is a blockchain platform that was specifically designed for use in enterprise systems, and it possesses a set of features that can be useful in distributed robotic systems. These features include: the identification of participants for secure identity management and certificate generation, permissioned networks for added data security, high performance for real-time robotic data processing, configurable low-latency transaction confirmation for real-time consensus, and the ability to partition data channels and maintain the privacy and confidentiality of transactions, which can seamlessly integrate with the pub/sub system of ROS\,2.

\begin{figure*}[t]
    \centering
    \includegraphics[width=.9\textwidth]{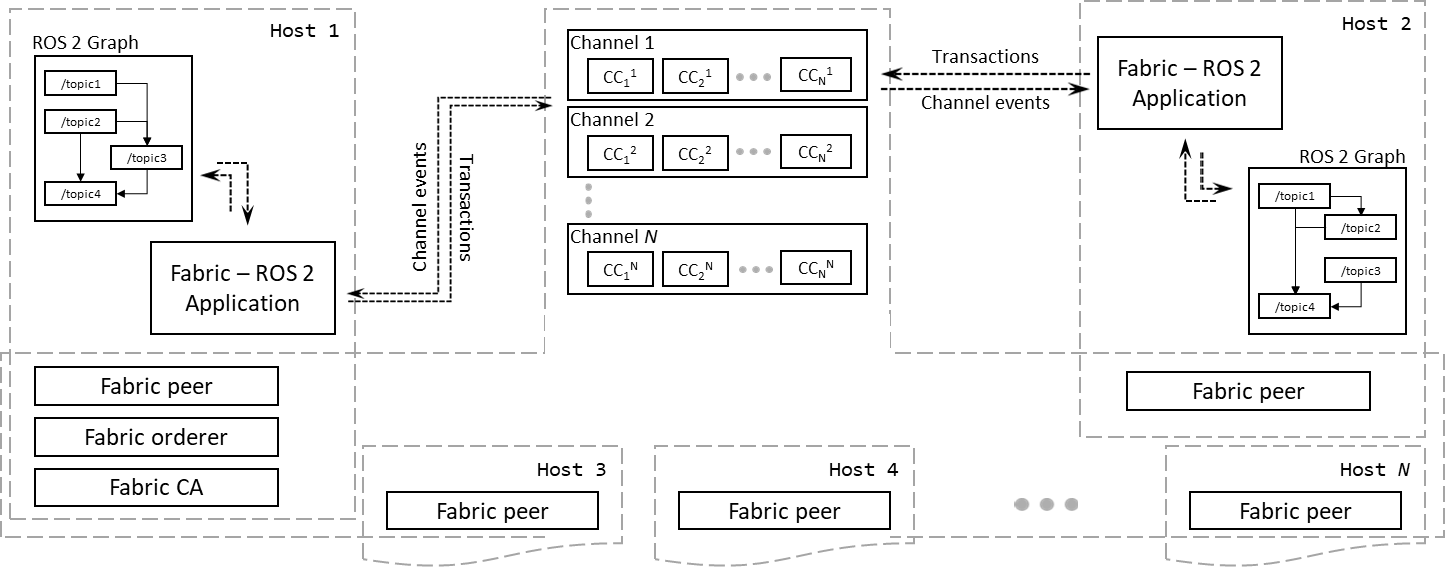}
    \caption{System architecture overview with and illustration of the ROS\,2 and Fabric blockchain interfaces.}
    \label{fig:architecture}
    \vspace{-2em}
\end{figure*}

A framework for controlling robots and gathering and processing data using blockchain technology by integrating ROS\,2 and Hyperledger Fabric blockchain is examined in~\cite{salimi2022towards} and~\cite{salimi2022secure}. In a demonstration, robots were used to detect objects of interest for inventory management, and it was found that the blockchain layer did not significantly affect robot performance. \cite{salimi2022secure} builds upon this framework by incorporating ultra-wideband (UWB) localization for autonomous navigation and robot collaboration, specifically in an inventory management application within a warehouse-like environment. And~\cite{torrico2022UWBRole} presents a way of achieving distributed decision-making in a system of multiple mobile robots with the help of dynamic UWB role allocation algorithms integrated into Fabric smart contracts. It presents the same functionality in a secure and trustworthy way, the enhancement of identity and data access management.

%%%%%%%%%%%%%%%%%%%%%%%%%%%%%%%%%%%%%%%%%%%%%%
%%                                          %%
%%        PROBLEM DEFINITION                %%
%%                                          %%
%%%%%%%%%%%%%%%%%%%%%%%%%%%%%%%%%%%%%%%%%%%%%%

% \newpage
\section{System Architecture}

We now describe the integration approach we follow for Fabric and ROS\,2 with an event-based architecture. We have redesigned our approach from previous works from the ground up, with a novel design that is more parametrizable, more scalable and has wider flexibility for in-chain or off-chain data confirmation.

\subsection{Fabric Network Configuration}

A set of organizations hosts the Fabric network, represented by a series of peer nodes. They are generally containers for the certificate authorities (CA) and  peers to verify their membership in the network and are also called members of the network. CAs provide cryptographic signatures for all operations performed inside Hyperledger Fabric. In addition to generating the necessary certificates for the nodes, they define the organization's definitions and applications. As trusted identifiers of components belonging to specific organizations, CAs are vital in the network.

Fabric networks are distinguished from other blockchain solutions by their private data channels through which the network can be partitioned and a global ledger maintained while partitioning. Channels are private subnets that allow two or more network members to communicate securely. Smart contracts are supported by chaincodes deployed within channels. Organizations use chaincode definitions to agree on the parameters of a chaincode before it can be used on a channel. In order to use the chaincode, each channel member must approve its definition. Upon approval by enough channel members, chaincode definitions can be committed to the channels and then when the chaincode is invoked for the first time, its state will be committed to the corresponding channel.

Lastly, smart contracts define a set of functions that peers from different organizations must implement before they can transact and they contain agreements that specify common terms, data, rules, concept definitions, and processes. An application external to the blockchain usually invokes smart contracts  and provides an interface for interacting with the ledger. A general-purpose programming language such as Go, Java or NodeJS can be used to write both applications and smart contracts.

\subsection{Architecture Overview}

The overall system architecture is illustrated in Fig.~\ref{fig:architecture}. The Fabric network effectively acts as the data transport layer between individual ROS\,2 subsystems. ROS\,2 is becoming the \textit{de-facto} standard in the next-generation of robotic systems. This new version of the Robot Operating System (ROS) middleware, has been designed based on the following key principles: (i) distribution, (ii) abstraction, (iii) asynchrony, and (iv) modularity~\cite{macenski2022robot}. ROS\,2 has therefore been designed as an inherently distributed framework, yet multiple challenges remain in terms of real-world deployments~\cite{zhang2022distributed}. The ROS\,2 communication layer leverages different available DDS protocol implementations. The out-of-the-box DDS configuration does not scale well even for a few robots, with the literature showing the need for more performant solutions such as Zenoh for inter-robot communication~\cite{xianjia2023loosely}. Despite this, ROS\,2 is still a preferred option as, even in the single-robot case, it eliminates the single-point-of-failure ROS master in ROS\,1, while providing a more modern API and functionality in some areas. DDS is indeed optimized for low-latency and high-throughput data exchange, but the majority of use cases to date are within a single organization or network. A Fabric network, on the other hand, is optimized for cross-organizational data exchange and secure, tamper-proof data storage and sharing.

In general words, we do not intend for the Fabric bridge presented in this paper to replace DDS even in cross-organizational networks, but we argue that it can provide significant advantages in terms of auditability, security and even stability for remote operations where devices are not in the same network. Both in our previous works~\cite{salimi2022towards, salimi2022secure, torrico2022UWBRole}, and in this paper, our experiments show that the Fabric network is able to handle hundreds of messages per second, which can suffice in many use cases for critical data including robot states (e.g., odometry and localization data in mobile robots, or joint states for manipulators) as well as control inputs, specially in teleoperation of critical assets.

The key difference with respect to previous works is the introduction of an event-driven architecture, where ROS\,2-Fabric applications execute callbacks from both the ROS\,2 and the Fabric networks they are interfacing. ROS\,2 events occur when, for example, new messages are published to a given topic. Based on the bridge application configuration, these messages can be forwarded to the Fabric network in the forms of transactions, e.g., the creation or modification of assets in different channel chaincodes. Fabric channel events occur when new transactions become confirmed in the blockchain.

\begin{figure}[!ht]
    \centering
    \includegraphics[width=.49\textwidth]{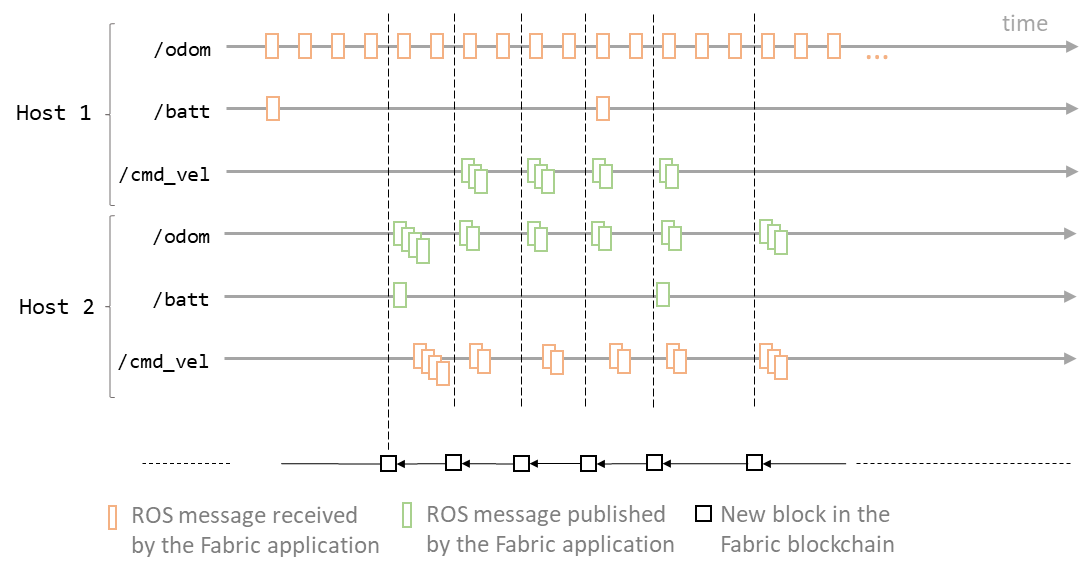}
    \caption{Overview of data transport latencies in a simplified use case. Messages are stacked for illustrative purposes when they overlap in time.}
    \label{fig:latencies_over_time}
\end{figure}

Because for transactions to be confirmed a new block has to be consensuated in the blockchain, there is an inherent latency for the data transport. In our previous work~\cite{salimi2022towards} we have studied the latency and computational load for different Fabric network configuration. In this paper, we show how the event-driven approach significantly reduces data transport latency at high network capacity, even if the confirmation time remains similar at lower network loads. An illustrative example of the timing of messages is shown in Fig.~\ref{fig:latencies_over_time}, where we only show a simple two-host system where odometry, battery status and velocity control data is transmitted through the blockchain. It is worth noticing that the transport latencies are non-deterministic; because applications in both hosts issue transactions in the Fabric network, it is not evident a priori which transactions will be included in a given block in the blockchain.

The architecture diagram in Fig.~\ref{fig:architecture} also shows that the logic in the blockchain layer can be structured in different channels and chaincodes. For a more detailed explanation of these concepts, we refer the reader to our previous work~\cite{salimi2022towards}.

%%%%%%%%%%%%%%%%%%%%%%%%%%%%%%%%%%%%%%%%%%%%%%
%%                                          %%
%%              METHODOLOGY                 %%
%%                                          %%
%%%%%%%%%%%%%%%%%%%%%%%%%%%%%%%%%%%%%%%%%%%%%%

% \newpage

\section{Methodology}
\label{sec:methodology}

In this section, we describe the experimental settings and implementation details.

\subsection{Fabric and ROS\,2 Integration}

In our previous work~\cite{salimi2022towards}, we show a proof-of-concept for ROS\,2 and Fabric integration using \textit{rclgo}\footnote{\url{https://github.com/tiiuae/rclgo}} to build applications using Go that are able to interact with both the Fabric blockchain and the ROS\,2 computational graph. However, there are a number of limitations in the previous design and implementation. First, The nature of Go requires a priori knowledge of data message types, from the ROS\,2 perspective, to be known to the applications deployed across the network. This limits the parametrization of applications and ease of adoption. Second, the fabric-sdk-go\footnote{\url{https://github.com/hyperledger/fabric-sdk-go}} interface to Fabric is less mature that other Fabric SDK implementations, such as NodeJS, even if performing better in some aspects~\cite{foschini2020hyperledger}. This leads, for example, to issues integrating channel events. Therefore, the implementation in the previous work relies on polling the state of the blockchain in order to monitor changes and transactions from other nodes in the network, which in turn leads to a performance bottleneck.

To address the issues listed above, we have redesigned the Fabric-ROS\,2 interface from the ground up, with a focus on modularity, parametrization and generalization, and scalability through an event-driven approach. We have used NodeJS, a back-end JavaScript runtime environment that can be utilized for developing applications on the Fabric blockchain. We are able to interact with the blockchain network through Fabric Node SDK\footnote{\url{https://github.com/hyperledger/fabric-sdk-node}}, enabling submission of transactions and performance of queries.

In the system proposed in this paper, we employ peer channel-based event services to receive chaincode events generated by successfully committed transactions. As an event-driven programming language, we can easily combine NodeJS and the chaincode events to trigger external processes in response to ledger updates. The client API provides a mechanism to handle these events using NodeJS. In our case, whenever a ROS\,2 message is published to one of the pre-configured topics, the Fabric application receives the data through \textit{rclnodejs}\footnote{\url{https://github.com/RobotWebTools/rclnodejs}}, the ROS\,2 NodeJS client library. The data origin can then be validated (e.g., ensuring only the Fabric peer in a given robot is able to send state estimation data concerning that same robot) and forwarded to the blockchain network as a transaction. %and upload to the Fabric network to share to other users. 
The integration of Fabric with ROS\,2 allows for ROS\,2 data transmission across different ROS\,2 networks, thereby enabling remote control that is both feasible and secure.

\begin{algorithm}[t]
    \footnotesize
    \SetAlgoLined
    \KwData{parameters}
    Contains:\\
	    \hspace{1em}Asset ID: /ros\_msg\_ID \\
	    \hspace{1em}Asset data: \\
	    \hspace{2em}\scriptsize{\tcp{ROS\,2 message example}}
	    \hspace{2em}topic: /ros\_topic\_name  \\
	    \hspace{2em}msg\_type: ros\,2\_msg\_type  \\
	    \hspace{2em}msg\_data: \{header: std_msgs/Header, \dots\}  \\
	\BlankLine
	\textbf{Application 1: ROS subscriber + Fabric publisher}\\
    \tcp{\scriptsize{Initialize Hyperledger Fabric client}}
    fabric\_client $\gets$ initFabricClient()\;
    
    \tcp{\scriptsize{Subscribe to desired ROS topics}}
    subscriber $\gets$ subscribeToTopics(parameters)\;
    
    \While{true}{
        \tcp{\scriptsize{Receive message from ROS~2 topic}}
        ros\_message $\gets$ subscriber.receive()\;
        
        \tcp{\scriptsize{Create asset in Hyperledger Fabric}} fabric\_client.createAsset(ros\_message.ID, asset\_data)\;
        \tcp{\scriptsize{Listen to channel events}}
        event $\gets$ fabric\_client.listenToChannelEvents()\;
        }
    \caption{Fabric subscriber}
    \label{alg:pseudo_algotirhm1}

\end{algorithm}
    
\begin{algorithm}[t]
    \footnotesize
    \SetAlgoLined
    \textbf{\footnotesize Application 2: Fabric subscriber + ROS publisher }\\
    \tcp{\scriptsize{Initialize Hyperledger Fabric client}}
    fabric\_client $\gets$ initFabricClient()\;
    subscriber $\gets$ subscribeToTopics(parameter);\\
        \If{event.isRelevant(parameters)}{
            \tcp{\scriptsize{Get asset data from event}}
            asset\_data $\gets$ fabric\_client.getAssetData(event.asset\_id)\;
            
            \If{asset\_data.topic\_name != subscriber}{
            
            \tcp{\scriptsize{Publish asset data as ROS~2 message}} 
            publisher.publish(asset\_data.topic\_name, asset\_data.msg\_type, asset\_data.msg\_data)\;
            }
        }
    \caption{Fabric subscriber}
    \label{alg:pseudo_algotirhm2}
\end{algorithm}

\subsection{Fabric Smart Contracts}

The smart contracts, or chaincodes, in this paper are implemented in Golang. We provide a generic and streamlined smart contract, compared with our previous works, with a generic data structure that can accommodate any type of ROS message. The smart contract functions are also simplified and reduced to the minimum: creation or edition of assets in a single \textit{set} method, query of single or all assets in a single \textit{get} method, and deletion of assets in addition to those. The key difference with respect to the previous implementation is that the methods now register channel events including the type and payload of transactions. This allows for applications to listen to events across chaincodes without requiring them to know a priori their structure or available methods.

\begin{figure}[t]
    \centering
    \includegraphics[width=.45\textwidth]{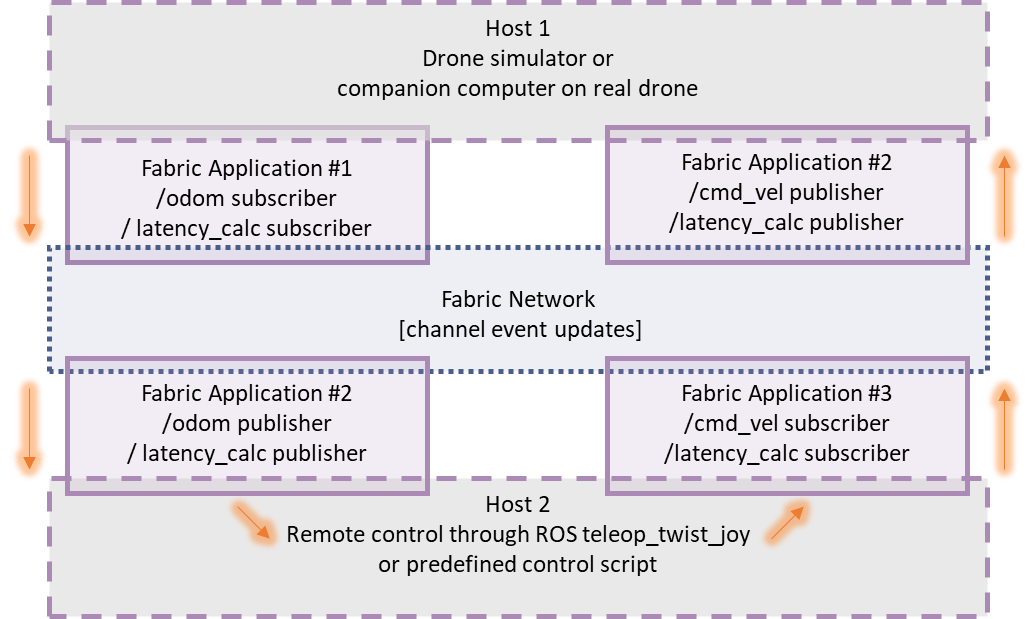}
    \caption{Overview of the implementation.}
    \label{fig:implementation_overview}
    
    \vspace{-2em}
\end{figure}

\subsection{ROS Client Applications}

Fabric ROS\,2 client applications are specifically designed to interact with both the Hyperledger Fabric blockchain network and the ROS\,2 computational graph of a system. For each hosts, at least two applications, namely \textit{Fabric publisher} and \textit{Fabric subscriber} have been deployed to deal with the assets. The \textit{Fabric publisher} forwards data from ROS\,2 to Fabric while the \textit{Fabric subscriber} operates in the opposite direction. The two applications can be combined in a single one, but we decide to divide the logic for easier parametrization and configuration.

To provide a detailed description of the application's functionality, we summarize the key operations in Algorithm~\ref{alg:pseudo_algotirhm1} and Algorithm~\ref{alg:pseudo_algotirhm2}.

\subsection{Experimental settings}

In order to assess the performance of the proposed approach, we perform a series of simulation and real-world experiments. The objective of the experiments is to show (i) superior performance with respect to previous solutions; and (ii) viability of new teleoperation use cases that are made possible with the event-driven interface.

\subsubsection{Hardware}

The Fabric network is set up with three different computers, with various hardware capabilities. In all cases, the computational load of running the Fabric blockchain remains vastly negligible, as we have shown in our previous work~\cite{salimi2022towards}. The three hosts have Intel i7-9750H, i7-12700 and Intel i3-1215U processors, and 64\,GB, 32\,GB and 16\,GB of memory, respectively.

\begin{figure}[b]
    \centering
    \includegraphics[width=0.42\textwidth]{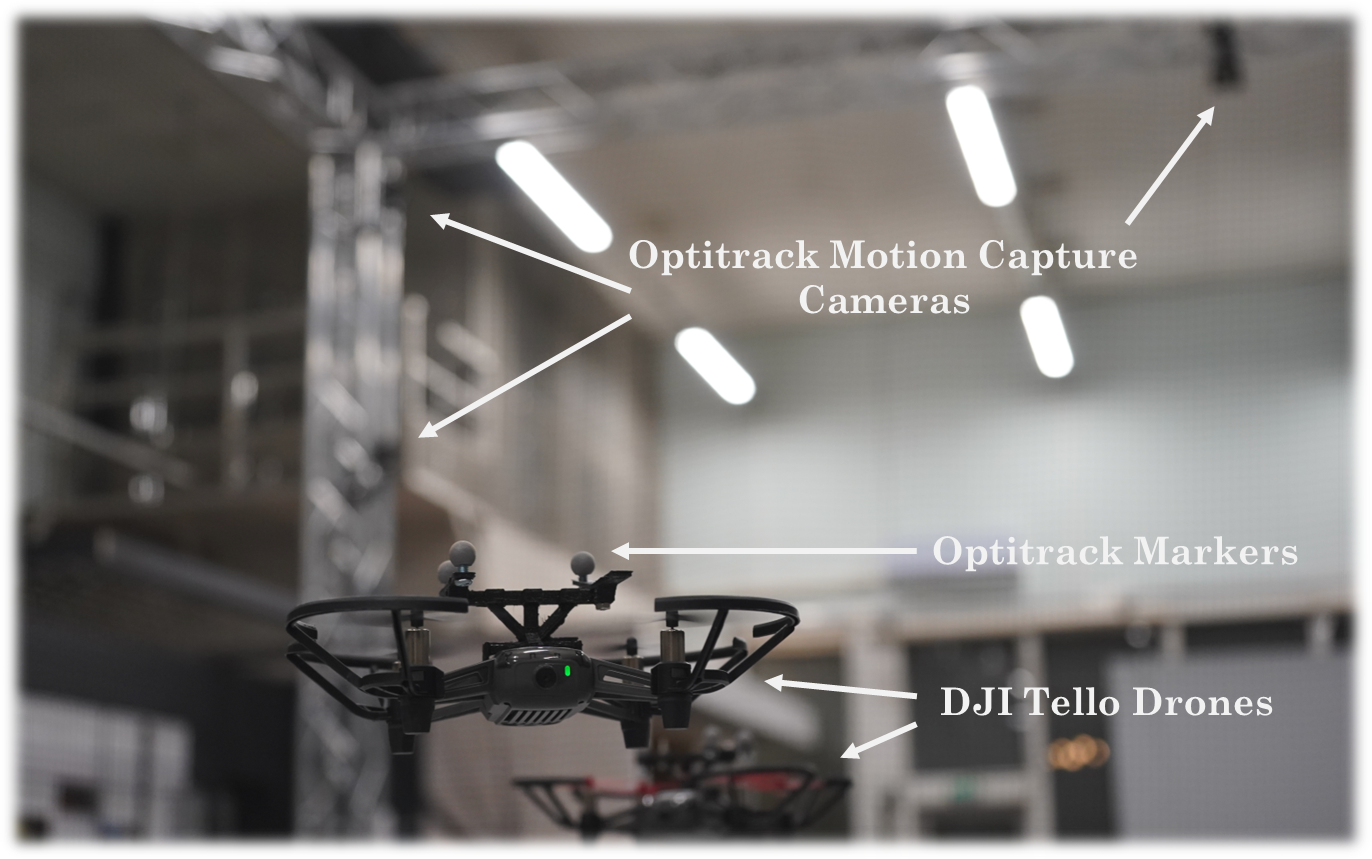}
    \caption{Tello drones in the Optitrack motion capture system.}
    \label{fig:sexy_tellos}
\end{figure}

% We used 
For the real-world experiments, we use a commercially available Ryze Tello MAV. The drone is equipped with Optitrack markers for localization, as shown in Fig.~\ref{fig:sexy_tellos}.

\subsubsection{Simulation environment}

In addition to the experiments with the Tello MAVs, we also simulate larger-scale environments to assess the scalability of the proposed approach. We use a ROS\,2 Gazebo simulation where we configure multiple drones\footnote{\url{https://github.com/TIERS/tello-ros2-gazebo}}.

\subsubsection{Software}

Robot control and localization are running in Host\,1. Data from the Optitrack MOCAP system is received with a VRPN client ROS node and used for accurate position and attitude of the robot. The teleoperation application and a basic motion controller are running in Host\,2. Both hosts run ROS\,2 Galactic under Ubuntu\,20.04. Two Fabric applications are then deployed in each host to interface with the ROS\,2 systems. The basic structure of the implementation is shown in Fig.~\ref{fig:implementation_overview}. The applications acts as a ROS subscriber and Fabric publisher and vice-versa in both hosts. The full implementation will be made available in our GitHub\footnote{\url{htpps://github.com/tiers/ros2fabric}}.
\subsection{Metrics}

In this paper, we aim to evaluate the feasibility of near real-time robot control through the proposed Fabric bridge. To analyze the performance of the proposed system, we collect and analyze different metrics. First, we measure the two-way latency and compare it to a baseline where the data in the blockchain is regularly polled, following the implementation in the previous works. Second, we show that the proposed system is usable in both simulation and real-world experiments, with different controllers while measuring the two-way data transport latency. Third, we push the limits of the standard trajectory controllers to study at what point the transport latency between the Fabric nodes is too large to enable effective control.

%%%%%%%%%%%%%%%%%%%%%%%%%%%%%%%%%%%%%%%%%%%%%%
%%                                          %%
%%              EXPERIMENTS                 %%
%%                                          %%
%%%%%%%%%%%%%%%%%%%%%%%%%%%%%%%%%%%%%%%%%%%%%%

% \newpage
\section{Experimental Results}

In this section we show experimental results from simulators and real-world experiments, to show the applicability of the proposed interface for remote control of ROS\,2 robots through a Fabric blockchain and the improvements with respect to the previous solutions.

\begin{figure}
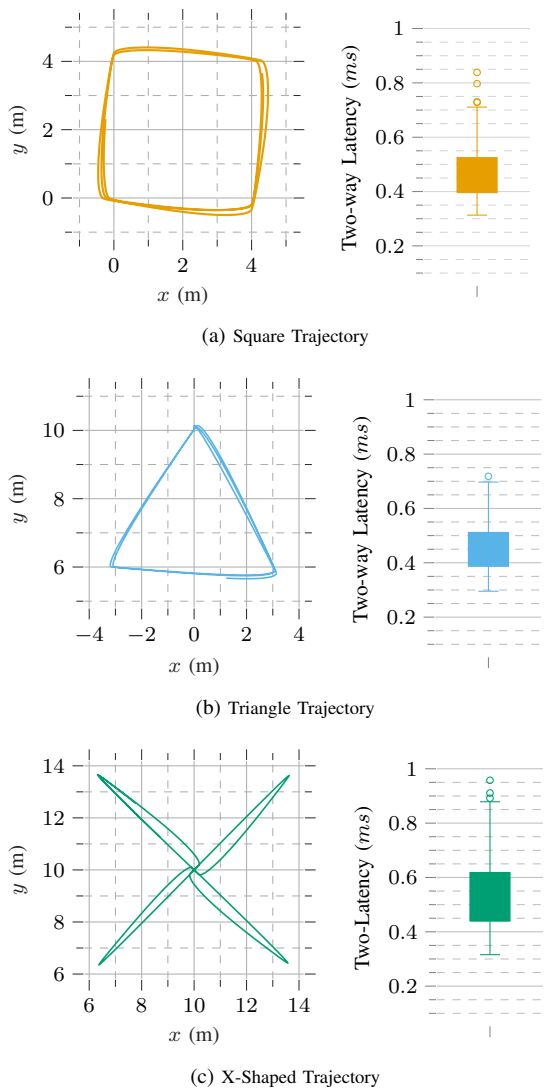

    \centering

    \begin{subfigure}{.42\textwidth}
        \setlength{\figurewidth}{.6\textwidth}
        \setlength{\figureheight}{.6\textwidth}
        \footnotesize{\input{tex/single_drone_sims/batch5Msg10Square_trajectory.tex}}
        \setlength{\figurewidth}{.4\textwidth}
        \setlength{\figureheight}{.7\textwidth}
        \footnotesize{% This file was created with tikzplotlib v0.10.1.
\begin{tikzpicture}

\definecolor{darkgray148163195}{RGB}{148,163,195}
\definecolor{darksalmon232149117}{RGB}{232,149,117}
\definecolor{darkslategray38}{RGB}{38,38,38}
\definecolor{black}{RGB}{89,89,89}
\definecolor{green}{RGB}{0,128,0}
\definecolor{lightgray204}{RGB}{204,204,204}
\definecolor{mediumaquamarine113182160}{RGB}{113,182,160}

\definecolor{darkgray176}{RGB}{176,176,176}
\definecolor{darkorange25512714}{RGB}{255,127,14}

% \definecolor{color0}{rgb}{0.1,0.1,0.1}
% \definecolor{color1}{rgb}{0.83921568627451,0.152941176470588,0.156862745098039}
% \definecolor{color2}{rgb}{0.12156862745098,0.466666666666667,0.705882352941177}
% \definecolor{color3}{rgb}{0.580392156862745,0.403921568627451,0.741176470588235}
% \definecolor{color4}{rgb}{0.890196078431372,0.466666666666667,0.76078431372549}

\definecolor{color1}{rgb}{0.90, 0.62, 0.00}
\definecolor{color2}{rgb}{0.34, 0.70, 0.91}
\definecolor{color3}{rgb}{0.00, 0.62, 0.45}
\definecolor{color4}{rgb}{0.94, 0.89, 0.27}
\definecolor{color5}{rgb}{0.00, 0.45, 0.69}
\definecolor{color6}{rgb}{0.83, 0.37, 0.00}

\begin{axis}[
    width=\figurewidth,
    height=\figureheight,
    axis line style={white},
    tick align=outside,
    tick pos=left,
    xtick={1},
    xticklabels={},
    ylabel={Two-way Latency ($ms$)},
    minor y tick num = 3,
    minor y grid style={dashed},
    ymajorgrids,
    ymajorgrids=true,
    ymajorticks=true,
    yminorgrids,
    yminorgrids=true,
    enlarge y limits=.05,
    %
    %
    %
    %
    % tick align=outside,
    % tick pos=left,
    % x grid style={darkgray176},
    xmin=0.8, xmax=1.2,
    % xtick style={color=black},
    % y grid style={darkgray176},
    ymin=0.1, ymax=1.02,
    % ytick style={color=black}
]
\addplot [draw=color1,  fill=color1]
table {%
0.925 0.3961391755
1.075 0.3961391755
1.075 0.524946124
0.925 0.524946124
0.925 0.3961391755
};
\addplot [draw=color1,  fill=color1]
table {%
1 0.3961391755
1 0.313193731
};
\addplot [draw=color1,  fill=color1]
table {%
1 0.524946124
1 0.710816265
};
\addplot [draw=color1,  fill=color1]
table {%
0.9625 0.313193731
1.0375 0.313193731
};
\addplot [draw=color1,  fill=color1]
table {%
0.9625 0.710816265
1.0375 0.710816265
};
\addplot [draw=color1,  fill=color1, mark=o, mark size=1.23, mark options={solid,fill opacity=0}, only marks]
table {%
1 0.727717708
1 0.79721908
1 0.730365483
1 0.838216028
};
\addplot [draw=color1,  fill=color1]
table {%
0.925 0.45569713
1.075 0.45569713
};
\end{axis}

\end{tikzpicture}}
        \caption{Square Trajectory}
        \label{subfig:sim_square_trajectory}
    \end{subfigure}

    \vspace{1em}

    \begin{subfigure}{.42\textwidth}
        \setlength{\figurewidth}{.6\textwidth}
        \setlength{\figureheight}{.6\textwidth}
        \footnotesize{\input{tex/single_drone_sims/batch5Msg10Triangle_trajectory.tex}}
        \setlength{\figurewidth}{.4\textwidth}
        \setlength{\figureheight}{.7\textwidth}
        \footnotesize{% This file was created with tikzplotlib v0.10.1.
\begin{tikzpicture}

\definecolor{color1}{rgb}{0.90, 0.62, 0.00}
\definecolor{color2}{rgb}{0.34, 0.70, 0.91}
\definecolor{color3}{rgb}{0.00, 0.62, 0.45}
\definecolor{color4}{rgb}{0.94, 0.89, 0.27}
\definecolor{color5}{rgb}{0.00, 0.45, 0.69}
\definecolor{color6}{rgb}{0.83, 0.37, 0.00}

\begin{axis}[
    width=\figurewidth,
    height=\figureheight,
    axis line style={white},
    tick align=outside,
    tick pos=left,
    xtick={1},
    xticklabels={},
    ylabel={Two-way Latency ($ms$)},
    minor y tick num = 3,
    minor y grid style={dashed},
    ymajorgrids,
    ymajorgrids=true,
    ymajorticks=true,
    yminorgrids,
    yminorgrids=true,
    enlarge y limits=.05,
    %
    %
    %
    %
    % tick align=outside,
    % tick pos=left,
    % x grid style={darkgray176},
    xmin=0.8, xmax=1.2,
    % xtick style={color=black},
    % y grid style={darkgray176},
    ymin=0.1, ymax=1.02,
    % ytick style={color=black}
]
\addplot [draw=color2,  fill=color2]
table {%
0.925 0.38701898575
1.075 0.38701898575
1.075 0.51173402525
0.925 0.51173402525
0.925 0.38701898575
};
\addplot [draw=color2,  fill=color2]
table {%
1 0.38701898575
1 0.295289985
};
\addplot [draw=color2,  fill=color2]
table {%
1 0.51173402525
1 0.697093246
};
\addplot [draw=color2,  fill=color2]
table {%
0.9625 0.295289985
1.0375 0.295289985
};
\addplot [draw=color2,  fill=color2]
table {%
0.9625 0.697093246
1.0375 0.697093246
};
\addplot [draw=color2,  fill=color2, mark=o, mark size=1.23, mark options={solid,fill opacity=0}, only marks]
table {%
1 0.718079433
};
\addplot [draw=color2,  fill=color2]
table {%
0.925 0.457784352
1.075 0.457784352
};
\end{axis}

\end{tikzpicture}}
        \caption{Triangle Trajectory}
        \label{subfig:sim_triangle_trajectory}
    \end{subfigure}

    \vspace{1em}

    \begin{subfigure}{.42\textwidth}
        \setlength{\figurewidth}{.6\textwidth}
        \setlength{\figureheight}{.6\textwidth}
        \footnotesize{\input{tex/single_drone_sims/batch5Msg10Xshape_trajectory.tex}}
        \setlength{\figurewidth}{.4\textwidth}
        \setlength{\figureheight}{.7\textwidth}
        \footnotesize{% This file was created with tikzplotlib v0.10.1.
\begin{tikzpicture}

\definecolor{color1}{rgb}{0.90, 0.62, 0.00}
\definecolor{color2}{rgb}{0.34, 0.70, 0.91}
\definecolor{color3}{rgb}{0.00, 0.62, 0.45}
\definecolor{color4}{rgb}{0.94, 0.89, 0.27}
\definecolor{color5}{rgb}{0.00, 0.45, 0.69}
\definecolor{color6}{rgb}{0.83, 0.37, 0.00}

\begin{axis}[
    width=\figurewidth,
    height=\figureheight,
    axis line style={white},
    tick align=outside,
    tick pos=left,
    xtick={1},
    xticklabels={},
    ylabel={Two-Latency ($ms$)},
    minor y tick num = 3,
    minor y grid style={dashed},
    ymajorgrids,
    ymajorgrids=true,
    ymajorticks=true,
    yminorgrids,
    yminorgrids=true,
    enlarge y limits=.05,
    %
    %
    %
    %
    % tick align=outside,
    % tick pos=left,
    % x grid style={darkgray176},
    xmin=0.8, xmax=1.2,
    % xtick style={color=black},
    % y grid style={darkgray176},
    ymin=0.1, ymax=1.02,
    % ytick style={color=black}
]
\addplot [draw=color3,  fill=color3]
table {%
0.925 0.439332749
1.075 0.439332749
1.075 0.6183315655
0.925 0.6183315655
0.925 0.439332749
};
\addplot [draw=color3,  fill=color3]
table {%
1 0.439332749
1 0.31574749
};
\addplot [draw=color3,  fill=color3]
table {%
1 0.6183315655
1 0.87865129
};
\addplot [draw=color3,  fill=color3]
table {%
0.9625 0.31574749
1.0375 0.31574749
};
\addplot [draw=color3,  fill=color3]
table {%
0.9625 0.87865129
1.0375 0.87865129
};
\addplot [draw=color3,  fill=color3, mark=o, mark size=1.23, mark options={solid,fill opacity=0}, only marks]
table {%
1 0.957635069
1 0.891833449
1 0.910942769
};
\addplot [draw=color3,  fill=color3]
table {%
0.925 0.520895289
1.075 0.520895289
};
\end{axis}

\end{tikzpicture}}
        \caption{X-Shaped Trajectory}
        \label{subfig:sim_x_trajectory}
    \end{subfigure}
    
    \caption{Simulation results: trajectories and two-way latencies for three different controllers. The latency is calculated since a odometry message is generated in one host until the matching velocity control message arrives from another host, including both the physical network and the Fabric network latencies.}
    \label{fig:sim_single_drone_results}
    \vspace{-2em}
\end{figure}

\subsection{Simulation results}

To first assess the viability of the proposed approach and study the data transport latency, we experiment with different types of trajectories for remote controlled drones. We use the three-host Fabric network described in Section~\ref{sec:methodology}, with the Gazebo simulator running in one host and the controller running in another host. Figure~\ref{fig:sim_single_drone_results} shows the recorded trajectory and two-way latencies for three different controllers. The latency is consistent across controllers, with minimal differences and staying below 1\,s.

The two-way latency is measured by keeping original ROS\,2 timestamps in a copy of messages in the controller system, forwarding them back to the simulator system. In the simulator, we are then able to compare the original timestamp stored in the ROS\,2 message with the current simulation time. This latency includes the transport latency over the wireless network in both directions, in addition to the confirmation time of two transactions in the Fabric blockchain, one in each direction. Therefore, we can conclude that the one-way transport latency plus the blockchain confirmation latency consistently stays between 350\,ms and 400\,ms. However, this metric is non-deterministic, as described earlier in the manuscript, owing to the dependency on the number of transactions that need to be accumulated before a new block is created in the blockchain and the data in it confirmed.

\begin{table}[t]
    \centering
    \caption{Comparison of performance in the new event-driven approach and previous polling bridge from~\cite{salimi2022towards} for a squared trajectory.}
    \label{tab:comparison}
    \footnotesize{
        \begin{tabular}{@{}cccccc@{}}
        % \footnotesize
            \toprule
            \textbf{State freq.} & \textbf{Load} & \textbf{Bridge} & \textbf{Avg. lat.} & \textbf{Max speed} & \textbf{CPU} \\
            (Hz) & (msg/s) & \textbf{method} & (s) & (m/s) & (\%) \\
            \midrule
            \multirow{2}{*}{30} & \multirow{2}{*}{70} & events & \textbf{0.55} & 0.3 & \textbf{0.2 - 0.4\%} \\
                                &                          & polling & \textbf{0.55} & 0.3 & 3 - 7\% \\
            \multirow{2}{*}{50} & \multirow{2}{*}{110} & events & \textbf{0.55} & \textbf{0.35} & 2 - 2.5\% \\
                                &                          & polling & N/A & N/A & N/A \\
            \bottomrule
        \end{tabular}
    }
\end{table}

\begin{figure}[b]
    \centering
    \setlength{\figurewidth}{.48\textwidth}
    \setlength{\figureheight}{.16\textwidth}
    \footnotesize{% This file was created with tikzplotlib v0.10.1.
\begin{tikzpicture}

\definecolor{darkgray148163195}{RGB}{148,163,195}
\definecolor{darksalmon232149117}{RGB}{232,149,117}
\definecolor{darkslategray38}{RGB}{38,38,38}
\definecolor{black}{RGB}{89,89,89}
\definecolor{green}{RGB}{0,128,0}
\definecolor{lightgray204}{RGB}{204,204,204}
\definecolor{mediumaquamarine113182160}{RGB}{113,182,160}

\definecolor{darkgray176}{RGB}{176,176,176}
\definecolor{darkorange25512714}{RGB}{255,127,14}

% \definecolor{color0}{rgb}{0.1,0.1,0.1}
% \definecolor{color1}{rgb}{0.83921568627451,0.152941176470588,0.156862745098039}
% \definecolor{color2}{rgb}{0.12156862745098,0.466666666666667,0.705882352941177}
% \definecolor{color3}{rgb}{0.580392156862745,0.403921568627451,0.741176470588235}
% \definecolor{color4}{rgb}{0.890196078431372,0.466666666666667,0.76078431372549}

\definecolor{color1}{rgb}{0.90, 0.62, 0.00}
\definecolor{color2}{rgb}{0.34, 0.70, 0.91}
\definecolor{color3}{rgb}{0.00, 0.62, 0.45}
\definecolor{color4}{rgb}{0.94, 0.89, 0.27}
\definecolor{color5}{rgb}{0.00, 0.45, 0.69}
\definecolor{color6}{rgb}{0.83, 0.37, 0.00}

\begin{axis}[
    width=\figurewidth,
    height=\figureheight,
    axis line style={white},
    tick align=outside,
    tick pos=left,
    ytick={1},
    yticklabels={},
    ylabel={Lat. ($ms$)},
    minor y tick num = 3,
    minor y grid style={dashed},
    ymajorgrids,
    ymajorgrids=true,
    ymajorticks=true,
    yminorgrids,
    yminorgrids=true,
    enlarge y limits=.05,
    %
    %
    %
    %
    % tick align=outside,
    % tick pos=left,
    % x grid style={darkgray176},
    xmin=0.4, xmax=1.3,
    % xtick style={color=black},
    % y grid style={darkgray176},
    ymin=0.85, ymax=1.2,
    % ytick style={color=black}
]
\addplot [draw=color6, fill=color6]
table {%
0.730587944 0.925
0.730587944 1.075
0.872862035 1.075
0.872862035 0.925
0.730587944 0.925
};
\addplot [draw=color6, fill=color6]
table {%
0.730587944 1
0.529246527 1
};
\addplot [draw=color6, fill=color6]
table {%
0.872862035 1
1.080564178 1
};
\addplot [draw=color6, fill=color6]
table {%
0.529246527 0.9625
0.529246527 1.0375
};
\addplot [draw=color6, fill=color6]
table {%
1.080564178 0.9625
1.080564178 1.0375
};
\addplot [draw=color6, fill=color6, mark=o, mark size=1.23, mark options={solid,fill opacity=0}, only marks]
table {%
0.493298911 1
0.457532281 1
0.498438912 1
0.500429283 1
1.09157994 1
1.095317609 1
1.109226344 1
1.216818089 1
};
\addplot [draw=color6, fill=color6]
table {%
0.801764586 0.925
0.801764586 1.075
};
\end{axis}

\end{tikzpicture}}
    \caption{Latency distribution for a simulation of three drones following the three trajectories in Fig.~\ref{fig:sim_single_drone_results}.}
    \label{fig:three_drones}
\end{figure}
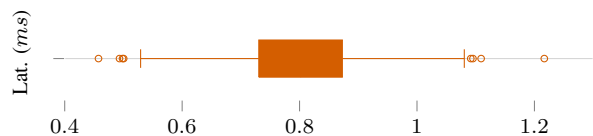

Additionally, we compare out approach to the previous polling-based implementation. We have optimized the previous solution presented in~\cite{salimi2022towards} and rewritten it in NodeJS for more fair comparison. The performance is shown in Table~\ref{tab:comparison}. While the one-way latency in the previous paper was reported to be near 1\,s, the current optimized solution is in par with the event-driven bridge. The main implementation difference is that a \textit{cache} of transaction metadata is kept at the application level. This leads to a more efficient polling process where only the most recent transactions need to be polled. It is worth noting that no actual transaction data is cached, only metadata. Despite the similar latency, the polling bridge falls short as soon as we increase the frequency of publication of robot state messages. As shown in the last row in Table~\ref{tab:comparison}, the polling fails completely when the total amount of messages surpasses 100\,msg/s, leading to diverging control latency.

\begin{figure}
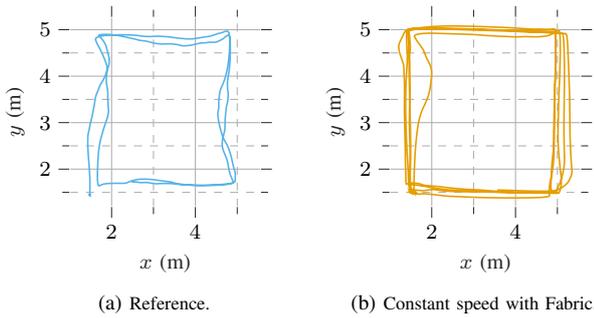

    \centering
    \begin{subfigure}{0.23\textwidth}
        \setlength{\figurewidth}{\textwidth}
        \setlength{\figureheight}{\textwidth}
        \footnotesize{\input{tex/real/good_reference_trajectory_no_fabric_trajectory.tex}}
        \caption{Reference.}
    \end{subfigure}
    \begin{subfigure}{0.23\textwidth}
        \setlength{\figurewidth}{\textwidth}
        \setlength{\figureheight}{\textwidth}
        \footnotesize{\input{tex/real/good_constant_with_fabric_trajectory.tex}}
        \caption{Constant speed with Fabric.}
    \end{subfigure}
    \caption{Experimental results from real-world experiments. Subfigure (a) shows a reference trajectory with a constant 0.5\,m/s speed for the Tello drone without the Fabric network. Subfigure (b) shows the trajectory with Fabric at the same constant speed.}
    \label{fig:real_references}
    \vspace{-2em}
\end{figure}

\begin{figure}
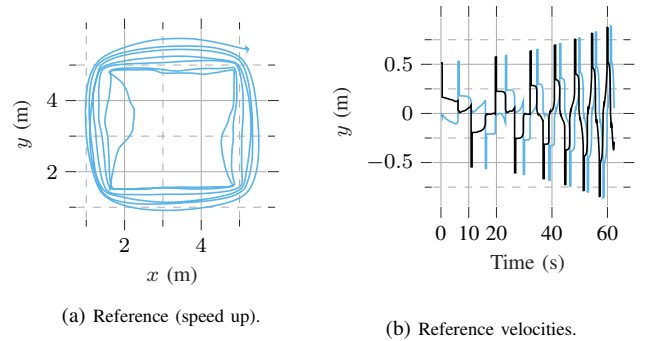
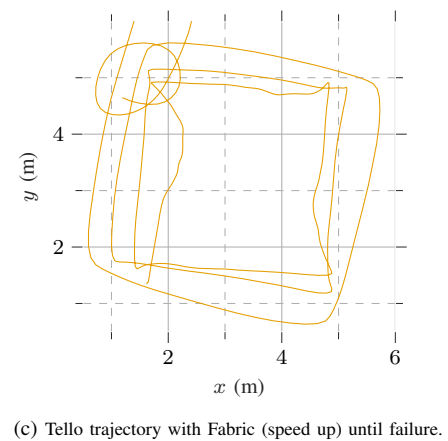

    \centering
    \begin{subfigure}{0.23\textwidth}
        \setlength{\figurewidth}{\textwidth}
        \setlength{\figureheight}{\textwidth}
        \footnotesize{\input{tex/real/good_reference_speed_no_fabric_trajectory.tex}}
        \caption{Reference (speed up).}
    \end{subfigure}
    \begin{subfigure}{0.23\textwidth}
        \setlength{\figurewidth}{\textwidth}
        \setlength{\figureheight}{\textwidth}
        \footnotesize{\input{tex/real/good_reference_speed_no_fabric_velocities.tex}}
        \caption{Reference velocities.}
    \end{subfigure}

    \vspace{1em}
    
    \begin{subfigure}{0.32\textwidth}
        \setlength{\figurewidth}{\textwidth}
        \setlength{\figureheight}{\textwidth}
        \footnotesize{\input{tex/real/good_speed_with_fabric_trajectory.tex}}
        \caption{Tello trajectory with Fabric (speed up) until failure.}
    \end{subfigure}

    \begin{subfigure}{0.42\textwidth}
        \setlength{\figurewidth}{\textwidth}
        \setlength{\figureheight}{.5\textwidth}
        \footnotesize{\input{tex/real/good_speed_with_fabric_velocities_both.tex}}
        \caption{Increasing linear Tello velocities until controller failure.}
    \end{subfigure}%
    \caption{\textit{Pushing the limits.} Example of failures of the controller when the two-way bridge latency is too large for the basic controller. Subfigure (b) is again a reference without Fabric but now increasing the linear speed at every turn. Subfigure (c) shows the x (black) and y (blue) components of the linear speed over time. Subfigure(c) shows the real-world trajectory when speed increases, while Subfigure (d) shows the velocity profiles until the controller fails. The \textit{virtual} velocities show the control input that would be generated if the controller was running in the same host as the Tello driver.}
    \label{fig:real_drone_failure}
    \vspace{-2em}
\end{figure}

Finally, to ensure the bridge stays performant with a higher number of drones, we show in Fig.~\ref{fig:three_drones} the latency distribution for a simulation with all three drone trajectories in Fig.~\ref{fig:sim_single_drone_results} simultaneously controlled, with increased robot state publication frequency. In this simulation, the total network load was over 250\,msg/sec (pushed to the network). This shows potential for scalability, which can be extended by separating the chaincodes across channels where subsets of network nodes participate.

\subsection{Real-world results}

In terms of the experiments with the Tello drones, we study the ability of the bridge to adapt to more realistic conditions, as the drone itself is controlled via a wireless interface from the host that was previously running the simulator. Additionally, the state estimation is now external from an Optitrack MOCAP system. These two changes add uncertainty and additional non-deterministic network latencies.

To show that the proposed system simply works out of the box, Fig.~\ref{fig:real_references} shows reference trajectories with and without the Fabric bridge for the Tello drone in the MOCAP arena. We then try to push the limits of the bridge, to showcase that there is an inherent trade-off between the controller complexity and the system performance due to the inherent network latencies. In Fig.~\ref{fig:real_drone_failure}, we experiment with increasing speeds. While a simple controller still performs well in simulation with linear velocities reaching 1\,m/s, we can observe that adding the Fabric bridge results in a controller failure as the robot state data used to generate a control input is too far in space from the actual robot position when the corresponding control is applied.

\subsection{Discussion}

Through this section, we have shown the real-world applicability of the proposed ROS\,2 - Fabric bridge for remote teleoperation of mobile robots. We have also compared the proposed approach to the previous implementation~\cite{salimi2022towards}, showing significant improvements in latency with the new event-driven design and implementation.

\subsubsection{Limitations:} the current work assumes a very simple controller that does not know the implications of the Fabric bridge, e.g., in terms of \textit{burst} publication of messages as they become available in a new blockchain block. At the same time, the bridge optimization we have carried out for larger number of robots assumes that the controller does not need all topic messages and is able to operate in the same way for an optimized subset of messages being published. From the robotics system perspective, a more optimal and network-aware controller might be able to stay performant even when the network becomes congested. Analogously, the current optimization does not necessarily apply to any control algorithm. However, our focus has been on studying and analysing the potential of the proposed ROS\,2-Fabric interface, and demonstrating its real-world applicability. It is therefore out of the scope of this paper how the control is actually carried out. We also intend for this to be used by human operators, yet we use a predefined and repeatable control program to be able to quantitatively assess the performance across different experimental settings.

\subsubsection{Future work potential:} this proof of concept has so far focused on pushing the limits of the Fabric blockchain for near real-time teleoperation. In addition to providing a more secure and auditable data transport later, from the point of view of functionality, there are other features from Fabric that we can leverage to augment the robotic system. First, we intend to combine the Fabric bridge with existing solutions to securing ROS\,2 systems, particularly SROS\,2~\cite{mayoral2022sros2, chen2022fogros}. This will allow us to provide more decentralized and cross-organizational identity management, private data separation in channels rather than individual topics, and the separation of higher-level logic (e.g., access control or conflicts between teleoperators) from the existing logic at the robot level. Second, there is potential to augment the teleoperation application with a estimate of the robot state based on real-time network latency monitoring and the knowledge of the data validation process through blocks in the Fabric blockchain. This would require a priori knowledge about the states that are being shared in the blockchain (e.g., odometry data in our experiments) as well as have a model of the system (e.g., a single integrator might suffice in the use case demonstrated in this paper). This augmentation of the teleoperation would theoretically enable the operator to produce a control command based on a likely robot state at the time that the control data will be delivered, in contrast to the time when the state description data was generated.
%%%%%%%%%%%%%%%%%%%%%%%%%%%%%%%%%%%%%%%%%%%%%%
%%                                          %%
%%              CONCLUSION                  %%
%%                                          %%
%%%%%%%%%%%%%%%%%%%%%%%%%%%%%%%%%%%%%%%%%%%%%%

% \newpage
\section{Conclusion}\label{sec:conclusion}

In this paper, we have proposed a new event-driven Fabric-ROS\,2 bridge to forward data between ROS\,2 systems. We have particularly focused on remote operations, with Fabric networks having advantages over ROS\,2  in terms of cross-organizational data transport. The current implementation of the proposed system significantly outperforms the performance of the previous works in the integration of ROS\,2 with Fabric blockchains. Our experimental results in both simulation and real-world settings show that two-way network latencies consistently stay below 0.5\,s even at relatively high network loads (over 100 messages/second for the robot state and control input). While these values are not comparable to the capacity of DDS within a single system, there are a number of security and data reliability advantages.

%%%%%%%%%%%%%%%%%%%%%%%%%%%%%%%%%%%%%%%%%%%%%%
%%                                          %%
%%            ACKNOWLEDGMENT                %%
%%                                          %%
%%%%%%%%%%%%%%%%%%%%%%%%%%%%%%%%%%%%%%%%%%%%%%

\section*{Acknowledgment}

This research work is supported by the Academy of Finland's RoboMesh (Grant No. 336061) and AeroPolis project (Grant No. 348480).

%%%%%%%%%%%%%%%%%%%%%%%%%%%%%%%%%%%%%%%%%%%%%%
%%                                          %%
%%              BIBLIOGRAPHY                %%
%%                                          %%
%%%%%%%%%%%%%%%%%%%%%%%%%%%%%%%%%%%%%%%%%%%%%%
% \newpage
% \nocite{*}
\bibliographystyle{unsrt}
\bibliography{bibliography}

\end{document}